\documentclass{article}

\usepackage{microtype}
\usepackage{graphicx}
\usepackage{subcaption}
\usepackage{booktabs} 

\usepackage{hyperref}

\usepackage[preprint]{icml2026}

\usepackage{amsmath}
\usepackage{amssymb}
\usepackage{mathtools}
\usepackage{amsthm}

\usepackage[capitalize,noabbrev]{cleveref}

\theoremstyle{plain}

\theoremstyle{definition}

\theoremstyle{remark}

\usepackage[textsize=tiny]{todonotes}

\icmltitlerunning{DualDiffusion: A Speculative Decoding Strategy for Masked Diffusion Models}

\begin{document}

\twocolumn[
  \icmltitle{DualDiffusion: A Speculative Decoding Strategy for Masked Diffusion Models}



  \icmlsetsymbol{equal}{*}

  \begin{icmlauthorlist}
  \icmlauthor{Satyam Goyal}{umich}
  \icmlauthor{Kushal Patel}{umich}
  \icmlauthor{Tanush Mittal}{umich}
  \icmlauthor{Arjun Laxman}{umich}
\end{icmlauthorlist}

\icmlaffiliation{umich}{Department of Computer Science and Engineering, University of Michigan, Ann Arbor, MI, USA}

\icmlcorrespondingauthor{Satyam Goyal}{sagoyal@umich.edu}

  \icmlkeywords{Machine Learning, ICML}

  \vskip 0.3in
]



\printAffiliationsAndNotice{\icmlEqualContribution}

\begin{abstract}
Masked Diffusion Models (MDMs) offer a promising alternative to autoregressive language models by enabling parallel token generation and bidirectional context modeling. However, their inference speed is significantly limited by the inability to cache key-value pairs due to bidirectional attention, requiring $O(N^2)$ computations at each generation step. While recent methods like FastDLLM and DkvCache improve inference speed through attention approximations and caching strategies, they achieve speedups at the cost of generation quality. We propose \textbf{DualDiffusion}, a speculative decoding framework for MDMs that combines fast drafter models (using efficient approximations) with slower, more accurate verifier models. By running multiple steps of a lightweight drafter followed by a single verification step, DualDiffusion achieves a superior Pareto frontier between generation steps and accuracy compared to existing approaches. We evaluate our method on MMLU and GSM8K, demonstrating that DualDiffusion maintains high accuracy while reducing the number of generation steps required, effectively pushing the quality-efficiency trade-off curve for masked diffusion language models.
\end{abstract}

\section{Introduction}

Masked Diffusion Models (MDMs) have emerged as a compelling alternative to autoregressive (AR) language models~\citep{sahoo2024simple,austin2021structured}. Unlike AR models that generate tokens sequentially from left to right, MDMs can unmask multiple tokens in parallel at each generation step while leveraging bidirectional attention over the entire sequence. This paradigm offers several advantages: faster generation through parallelization, improved controllability, and the ability to iteratively refine predictions using full sequence context.

Despite these benefits, MDMs face a critical computational bottleneck. The bidirectional attention mechanism that enables their superior context modeling prevents the use of key-value (KV) caching—a fundamental optimization in AR models. Consequently, each forward pass requires $O(N^2)$ attention computations over the full sequence length $N$, making inference significantly slower than cached AR generation, particularly for long sequences. This computational overhead limits the practical deployment of MDMs despite their theoretical advantages.

Recent work has attempted to address this efficiency gap through various approximation strategies. FastDLLM introduces caching approximations that fix previously generated tokens, while DkvCache employs dynamic cache management strategies. Esoteric Language Models~\citep{yang2025esoteric} explore hybrid approaches combining causal and bidirectional attention patterns. However, these methods fundamentally trade generation quality for speed: by constraining the attention mechanism or fixing token representations, they sacrifice the iterative refinement capability that makes MDMs powerful. Our work aims to resolve the quality tradeoff these methods have.

Speculative decoding has proven highly effective for accelerating AR models~\citep{leviathan2023fast,chen2023accelerating}. The key insight is to use a fast, approximate "drafter" model to propose multiple tokens, which are then verified by a slower, accurate model in a single forward pass. This preserves the output distribution of the accurate model while amortizing its computational cost. However, applying this technique to MDMs presents unique challenges: unlike AR models where token probabilities are conditionally independent given the prefix, MDM token generation is iterative, meaning that it requires multiple steps to achieve the same expected quality from a verifier. Standard verification strategies from AR speculative decoding cannot directly transfer.

We introduce \textbf{DualDiffusion}, a speculative decoding framework designed specifically for masked diffusion models. Our approach runs multiple unmasking steps using a fast drafter model (which may employ approximations like FastDLLM or DkvCache), then performs a single verification step with an accurate verifier model (e.g., LLaDA~\citep{sahoo2024simple}). We develop novel verification algorithms that probabilistically remask tokens based on confidence scores or divergence measures between drafter and verifier predictions. This allows us to recover from drafter errors while maintaining the iterative refinement benefits of MDMs.

Our contributions are as follows:
\begin{itemize}
\item We propose DualDiffusion, the first speculative decoding framework for masked diffusion language models, combining fast approximate drafters with accurate verifiers.
\item We introduce three verification algorithms—Trust Verifier, KL Divergence-based, and Confidence-based remasking—to selectively correct drafter predictions.
\item We demonstrate that DualDiffusion achieves a superior Pareto frontier on the accuracy-efficiency trade-off, maintaining high accuracy with fewer generation steps compared to baseline MDMs and existing acceleration methods.
\item We provide empirical validation on MMLU and GSM8K benchmarks, showing consistent improvements over LLaDA and FastDLLM baselines.
\end{itemize}

\section{Background}

\subsection{Masked Diffusion Models}

Masked Diffusion Models (MDMs) represent a prominent class of discrete diffusion models for text generation~\citep{sahoo2024simple,austin2021structured}. In MDMs, the forward process progressively masks tokens in a sequence, while the reverse process iteratively unmasks them to generate coherent text.

Given a sequence of tokens $\mathbf{x}_0 = (x_1, x_2, \ldots, x_L)$ of length $L$, the forward masking process at continuous time $t \in [0,1]$ is defined as:
\begin{equation}
q_t(x_i | x_{0,i}) = \text{Cat}(x_i; (1-\alpha_t)x_{0,i} + \alpha_t m)
\end{equation}
where $\alpha_t$ is a noise schedule that increases from 0 to 1, and $m$ denotes a special \texttt{[MASK]} token. At $t=0$, the sequence is clean ($\alpha_0 \approx 0$), while at $t=1$, all tokens are masked ($\alpha_1 \approx 1$).

The reverse denoising process learns to predict the original tokens given a masked sequence. A denoising model $p_\theta(\mathbf{x}_0 | \mathbf{x}_t)$ is trained to maximize the evidence lower bound (ELBO):
\begin{equation}
\mathcal{L}_{\text{MDM}} = \mathbb{E}_{t, \mathbf{x}_t} \left[ \sum_{i \in \mathcal{M}(\mathbf{x}_t)} -\log p_\theta(x_{0,i} | \mathbf{x}_t) \right]
\end{equation}
where $\mathcal{M}(\mathbf{x}_t)$ denotes the set of masked token positions in $\mathbf{x}_t$.

During generation, MDMs start from a fully masked sequence $\mathbf{x}_1$ and iteratively unmask tokens over $T$ steps. At each step $t$, the model predicts probabilities for masked positions, and a subset of high-confidence tokens are unmasked based on a selection algorithm. This process continues until all tokens are revealed.

\subsection{The KV-Caching Bottleneck in MDMs}

A fundamental efficiency advantage of autoregressive (AR) models is \textbf{key-value (KV) caching}. During sequential generation, AR models cache the key and value tensors $\mathbf{K}^{[1:t-1]}$ and $\mathbf{V}^{[1:t-1]}$ from previously generated tokens. At step $t$, only the new token's query $\mathbf{Q}^{[t]}$ needs to be computed:
\begin{equation}
\mathbf{z}_t = \text{softmax}\left(\frac{\mathbf{Q}^{[t]} (\mathbf{K}^{[1:t]})^\top}{\sqrt{d_k}}\right) \mathbf{V}^{[1:t]}
\end{equation}
where $\mathbf{K}^{[1:t]} = [\mathbf{K}^{[1:t-1]}; \mathbf{K}^{[t]}]$ and $\mathbf{V}^{[1:t]} = [\mathbf{V}^{[1:t-1]}; \mathbf{V}^{[t]}]$. This reduces per-step complexity from $O(L^2)$ to $O(L)$.

However, MDMs employ \textbf{bidirectional attention}, where every token can attend to all positions in the sequence. This breaks two critical assumptions required for KV caching:

\paragraph{Timestep-variant representations.} In AR models with causal attention, the cached keys and values $\mathbf{K}^{[1:t-1]}$ and $\mathbf{V}^{[1:t-1]}$ remain fixed across all future steps. In MDMs, bidirectional attention means that the representation of a token at position $i$ can change at every denoising step $t$ as the context evolves. Formally, $\mathbf{K}_t^{[i]} \neq \mathbf{K}_{t'}^{[i]}$ for $t \neq t'$, preventing direct reuse of cached states.

\paragraph{Non-sequential decoding order.} AR models generate tokens strictly left-to-right, enabling deterministic computation of $\mathbf{Q}^{[t]}$, $\mathbf{K}^{[t]}$, $\mathbf{V}^{[t]}$ at position $t$. MDMs unmask tokens in arbitrary orders based on confidence scores or random sampling, making it impossible to predict which positions will be updated at each step.

These constraints force MDMs to recompute full attention over all $L$ tokens at every denoising step, resulting in $O(L^2)$ complexity per step and $O(TL^2)$ total complexity for $T$ steps, compared to $O(TL)$ for cached AR models.

\subsection{Existing Acceleration Methods}

Recent work has attempted to address MDM inference inefficiency through two main strategies: caching approximations and parallel decoding optimizations.

\paragraph{FastDLLM.} Fast-dLLM~\citep{wu2025fastdllm} introduces a block-wise approximate KV cache mechanism. The key insight is to divide the sequence into blocks and cache KV states from previously decoded blocks. Within each block, the cache is reused across multiple denoising steps and refreshed after block completion. Additionally, FastDLLM proposes confidence-aware parallel decoding, which dynamically selects tokens exceeding a confidence threshold for simultaneous unmasking rather than a fixed number per step. This mitigates degradation from the conditional independence assumption in parallel sampling. FastDLLM achieves up to 27.6$\times$ speedup with DualCache (caching both prefix and suffix blocks) but requires careful tuning of block size and threshold hyperparameters.

\paragraph{dKV-Cache.} dKV-Cache~\citep{ma2025dkvcache} observes that token representations in MDMs stabilize \emph{after} being decoded, not during their decoding step. This motivates a \textbf{delayed caching} strategy: KV states are cached one step after a token transitions from \texttt{[MASK]} to its decoded form. The method introduces two variants: (1) \textbf{dKV-Cache-Decode}, which maintains long-term caches with periodic refreshing to handle representation drift, and (2) \textbf{dKV-Cache-Greedy}, which restricts computation to a local window around the current decoding position, reducing complexity to $O(L^2)$ overall but with more aggressive cache eviction. Both variants achieve 2--10$\times$ speedup with minimal accuracy loss.

\paragraph{Limitations of existing methods.} While FastDLLM and dKV-Cache improve inference speed, they achieve these gains through architectural or caching approximations that inherently trade quality for efficiency. Block-wise caching in FastDLLM fixes token representations within blocks, preventing iterative refinement. dKV-Cache's delayed strategy and periodic refreshing introduce temporal inconsistencies. Both methods fundamentally constrain the bidirectional attention mechanism that makes MDMs powerful, resulting in a suboptimal Pareto frontier between generation quality and computational cost.

\subsection{Speculative Decoding in Autoregressive Models}

Speculative decoding has proven highly effective for accelerating AR model inference~\citep{leviathan2023fast,chen2023accelerating}. The core idea is to use a small, fast \textbf{drafter model} to propose multiple candidate tokens in parallel, which are then verified in a single forward pass by a larger, more accurate \textbf{verifier model}. Accepted tokens are retained, while rejected tokens are corrected. This preserves the output distribution of the verifier while amortizing its computational cost across multiple tokens.

The key guarantee in AR speculative decoding is that verification is deterministic: given a prefix, if the verifier would have generated token $x_i$ with probability $p_\theta(x_i | \mathbf{x}_{<i})$, and the drafter proposed $x_i$, then $x_i$ is accepted. This works because AR generation is conditionally independent given the prefix.

However, this guarantee does not directly transfer to MDMs. In masked diffusion, token probabilities are \emph{not} conditionally independent—the model simultaneously considers the entire bidirectional context. A verifier evaluating a drafter's proposals cannot simply check $p_\theta(x_i | \mathbf{x}_{\setminus i})$ because the context $\mathbf{x}_{\setminus i}$ itself may contain drafter errors that affect $x_i$. This necessitates novel verification strategies tailored to the non-autoregressive, context-dependent nature of MDMs.

\section{Methodology}

\subsection{DualDiffusion Pipeline Overview}

We propose DualDiffusion, a speculative decoding framework that combines fast drafter models with accurate verifier models to achieve superior quality-efficiency trade-offs for masked diffusion language models. The key insight is that while approximate acceleration methods like FastDLLM and dKV-Cache can generate candidates quickly, they sacrifice the iterative refinement capability that makes MDMs effective. By using these fast methods as drafters and correcting their errors with a high-quality verifier, we can maintain accuracy while reducing the total number of generation steps required.

The DualDiffusion pipeline operates in two phases:

\paragraph{Drafting Phase.} Starting from a masked sequence $\mathbf{x}_t$ at diffusion timestep $t$, we run $K$ unmasking steps using a fast drafter model $p_D$. The drafter may employ any acceleration technique—such as FastDLLM's block-wise KV cache or dKV-Cache's delayed caching—that trades quality for speed. After $K$ drafter steps, we obtain a partially unmasked sequence $\mathbf{x}_{t-K}^D$ where a subset of tokens have been decoded.

\paragraph{Verification Phase.} We perform a single forward pass using an accurate verifier model $p_V$ (e.g., LLaDA with full bidirectional attention) on the drafter's output $\mathbf{x}_{t-K}^D$. The verifier computes probability distributions $p_V(\cdot | \mathbf{x}_{t-K}^D)$ for all token positions. A verification algorithm then compares the drafter's predictions with the verifier's probabilities and selectively remasks tokens that are deemed unreliable. This corrected sequence $\mathbf{x}_{t-K}^V$ becomes the input for the next drafting phase.

By repeating this draft-verify cycle, DualDiffusion amortizes the cost of the expensive verifier across $K$ drafter steps while maintaining the quality of the full model. The effectiveness of this approach depends critically on the verification algorithm's ability to identify and correct drafter errors without unnecessary remasking.

\subsection{Verification Algorithms}

The verification phase must address a fundamental challenge: unlike AR speculative decoding, where token acceptance is deterministic given the prefix, MDM verification is inherently probabilistic due to bidirectional dependencies. We propose three verification strategies with increasing levels of sophistication.

\subsubsection{Trust Verifier}

The simplest approach is to accept the drafter's output without modification:
\begin{equation}
\mathbf{x}_{t-K}^V = \mathbf{x}_{t-K}^D
\end{equation}

This strategy provides maximum speedup by eliminating verification overhead, but offers no error correction. It is only effective when the drafter's quality is already high, such as when using dKV-Cache-Decode with frequent cache refreshing. The Trust Verifier serves as an ablation baseline to quantify the benefit of more sophisticated verification.

\subsubsection{KL Divergence-Based Remasking}

To detect tokens where the drafter and verifier disagree significantly, we compute the KL divergence between their predicted distributions at each unmasked position $i \in \mathcal{D}(\mathbf{x}_{t-K}^D)$, where $\mathcal{D}$ denotes the set of tokens decoded by the drafter:

\begin{equation}
D_{\text{KL}}^{(i)} = \text{KL}\big(p_D(\cdot | \mathbf{x}_{t-K}^D, i) \| p_V(\cdot | \mathbf{x}_{t-K}^D, i)\big)
\end{equation}

Tokens are remasked using one of two strategies:

\paragraph{Threshold-based remasking.} We remask position $i$ if $D_{\text{KL}}^{(i)} > \tau_{\text{KL}}$, where $\tau_{\text{KL}}$ is a hyperparameter. This produces a binary decision based on distribution mismatch.

\paragraph{Probabilistic remasking.} We remask position $i$ with probability proportional to the normalized divergence:
\begin{equation}
p_{\text{remask}}^{(i)} = \frac{D_{\text{KL}}^{(i)}}{\sum_{j \in \mathcal{D}} D_{\text{KL}}^{(j)}}
\end{equation}

This stochastic approach allows gradual error correction and can improve sample diversity.

The KL divergence criterion directly measures distributional disagreement between drafter and verifier, making it well-suited for identifying positions where the drafter's approximations (e.g., stale KV cache) have led to poor predictions.

\subsubsection{Confidence-Based Remasking}

An alternative approach exploits the verifier's confidence in its own predictions. For each unmasked position $i$, we compute the verifier's confidence as:
\begin{equation}
c_V^{(i)} = \max_{x \in \mathcal{V}} p_V(x | \mathbf{x}_{t-K}^D, i)
\end{equation}
where $\mathcal{V}$ is the vocabulary. Low confidence indicates that the verifier is uncertain even after observing the drafter's output, suggesting that position $i$ should be remasked for further refinement.

As with KL divergence, we support both threshold-based ($c_V^{(i)} < \tau_{\text{conf}}$) and probabilistic (remask with probability $1 - c_V^{(i)}$) remasking strategies.

Confidence-based remasking has the advantage of being unilateral—it depends only on the verifier's output, not on comparing drafter and verifier distributions. This makes it more robust when the drafter employs aggressive approximations that produce qualitatively different distributions.

\subsection{Algorithm}

Algorithm~\ref{alg:dualdiffusion} summarizes the complete DualDiffusion pipeline. The procedure alternates between drafting with $K$ steps of the fast model and verification with a single step of the accurate model. The number of drafter steps $K$ controls the speedup-quality trade-off: larger $K$ reduces verifier calls but may accumulate more drafter errors.

\begin{algorithm}[t]
\caption{DualDiffusion: Speculative Decoding for MDMs}
\label{alg:dualdiffusion}
\begin{algorithmic}[1]
\REQUIRE Drafter model $p_D$, verifier model $p_V$, sequence length $L$
\REQUIRE Drafter steps per cycle $K$, verification algorithm $\mathcal{A} \in \{\text{Trust}, \text{KL}, \text{Confidence}\}$
\STATE Initialize $\mathbf{x}_1 = [\texttt{MASK}]^L$ \COMMENT{Fully masked sequence}
\STATE $t \gets 1$ \COMMENT{Current diffusion timestep}
\WHILE{$\exists i : x_i = \texttt{MASK}$}
    \STATE \COMMENT{\textbf{Drafting Phase:} Run $K$ steps with fast drafter}
    \FOR{$k = 1$ to $K$}
        \STATE $\mathbf{x}_{t-k}^D \gets \text{Unmask}(p_D, \mathbf{x}_{t-k+1}^D)$ \COMMENT{Drafter unmasking step}
    \ENDFOR
    \STATE 
    \STATE \COMMENT{\textbf{Verification Phase:} Single step with accurate verifier}
    \STATE Compute $p_V(\cdot | \mathbf{x}_{t-K}^D)$ for all positions
    \STATE $\mathbf{x}_{t-K}^V \gets \mathcal{A}(p_D, p_V, \mathbf{x}_{t-K}^D)$ \COMMENT{Apply verification algorithm}
    \STATE 
    \STATE $\mathbf{x}_{t-K} \gets \mathbf{x}_{t-K}^V$ \COMMENT{Use verified sequence for next cycle}
    \STATE $t \gets t - K$
\ENDWHILE

$\mathbf{x}_0$ \COMMENT{Fully unmasked sequence}
\end{algorithmic}
\end{algorithm}

The choice of verification algorithm depends on the application requirements. Trust Verifier maximizes speed when drafter quality is acceptable. KL divergence-based remasking is effective when distributional alignment between drafter and verifier is critical. Confidence-based remasking is most robust when the drafter uses aggressive approximations that may produce overconfident but incorrect predictions.
\section{Experiments}

\subsection{Experimental Setup}

\paragraph{Datasets.} We evaluate DualDiffusion on two benchmarks: (1) \textbf{MMLU} (Massive Multitask Language Understanding)~\citep{hendrycks2021measuring} for assessing logical reasoning across diverse domains, and (2) \textbf{GSM8K}~\citep{cobbe2021training} for grade-school mathematical problem solving requiring multi-step arithmetic reasoning.

\paragraph{Models.} We use LLaDA~\citep{sahoo2024simple} as the verifier model and FastDLLM~\citep{wu2025fastdllm} as the drafter model. LLaDA employs full masked diffusion with bidirectional attention, while FastDLLM uses block-wise KV caching for acceleration. We compare DualDiffusion against: (1) \textbf{LLaDA} (verifier only, no acceleration), (2) \textbf{FastDLLM} (drafter only, maximum speedup), and (3) \textbf{DualDiffusion} with KL divergence-based verification (our method).

\paragraph{Metrics.} We report three metrics: (1) \textbf{Accuracy} on task-specific evaluation, (2) \textbf{Runtime} (seconds) for end-to-end generation, and (3) \textbf{Peak memory usage} (GB) during inference. All experiments are conducted on NVIDIA A40 GPUs with batch size 1.

\paragraph{Hyperparameters.} For DualDiffusion, we set the number of drafter steps $K=5$ and use KL divergence threshold $\tau_{\text{KL}} = 0.3$ for remasking. Additional ablations on $K$ and verification algorithms are provided in Section~\ref{sec:ablation}.

\subsection{Main Results}

\begin{table}[t]
\caption{Performance on MMLU benchmark.}
\label{tab:mmlu_results}
\vskip 0.15in
\begin{center}
\begin{small}
\begin{sc}
\begin{tabular}{lccc}
\toprule
Method & Accuracy & Time (s) & Memory (GB) \\
\midrule
FastDLLM      & 0.39 & 21.3  & 15.2 \\
LLaDA         & 0.48 & 320.5 & 16.8 \\
DualDiffusion & \textbf{0.47} & \textbf{82.0} & 31.7 \\
\midrule
\multicolumn{4}{l}{\textit{Speedup vs. LLaDA: 3.9$\times$}} \\
\bottomrule
\end{tabular}
\end{sc}
\end{small}
\end{center}
\vskip -0.1in
\end{table}

\begin{table}[t]
\caption{Performance on GSM8K benchmark.}
\label{tab:gsm8k_results}
\vskip 0.15in
\begin{center}
\begin{small}
\begin{sc}
\begin{tabular}{lccc}
\toprule
Method & Accuracy & Time (s) & Memory (GB) \\
\midrule
FastDLLM      & 0.21 & 18.7  & 15.2 \\
LLaDA         & 0.57 & 298.3 & 16.8 \\
DualDiffusion & 0.25 & \textbf{75.4} & 31.7 \\
\midrule
\multicolumn{4}{l}{\textit{Speedup vs. LLaDA: 4.0$\times$}} \\
\bottomrule
\end{tabular}
\end{sc}
\end{small}
\end{center}
\vskip -0.1in
\end{table}

\paragraph{MMLU Results.} DualDiffusion achieves 0.47 accuracy on MMLU, representing only a 0.01 degradation from the full LLaDA verifier (0.48) while delivering a 3.9$\times$ speedup (82.0s vs. 320.5s). This demonstrates that the KL divergence-based verification successfully corrects drafter errors in logical reasoning tasks. As expected, FastDLLM achieves the fastest runtime (21.3s) but suffers significant accuracy loss (0.39), confirming that aggressive caching approximations sacrifice quality. 

\begin{figure}[t]
  \begin{center}
    \centerline{\includegraphics[width=\columnwidth]{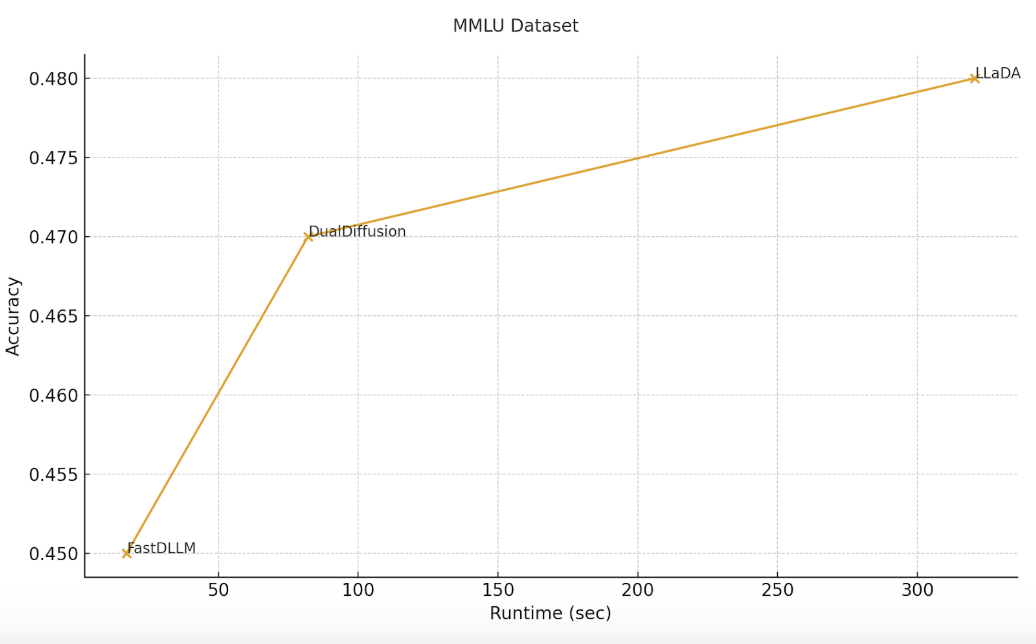}}
    \caption{Accuracy vs. runtime on MMLU. DualDiffusion (orange) achieves near-verifier accuracy at significantly reduced latency.}
    \label{fig:mmlu_accuracy}
  \end{center}
\end{figure}

Figure~\ref{fig:mmlu_accuracy} illustrates the accuracy-runtime Pareto frontier. DualDiffusion effectively bridges the gap between FastDLLM's speed and LLaDA's accuracy, achieving a superior trade-off compared to using either model alone.

\paragraph{Memory Overhead.} The primary cost of DualDiffusion is memory consumption. Maintaining both drafter and verifier weights in memory results in 31.7 GB peak usage, nearly double that of individual models ($\sim$16 GB). However, in deployment scenarios where GPU memory is abundant but latency is critical, this trade-off is favorable. Future work could explore model distillation or shared-parameter architectures to reduce this overhead.

\paragraph{GSM8K Results.} On GSM8K, DualDiffusion achieves 0.25 accuracy, substantially below the LLaDA verifier's 0.57 performance. This degradation reveals limitations in the current verification strategy for multi-step mathematical reasoning. Unlike MMLU's diverse logical questions, GSM8K requires precise sequential arithmetic where a single incorrect token can cascade into complete solution failure.

Analysis of the verification patterns indicates that the drafter frequently produces low-confidence predictions on arithmetic operations, but the KL divergence threshold ($\tau_{\text{KL}} = 0.3$) is insufficiently aggressive to remask these errors. When the drafter struggles on high-variance arithmetic tasks, the verifier's single correction step cannot recover the full solution path. This suggests that mathematical reasoning may benefit from: (1) more aggressive verification thresholds, (2) multi-step verification rather than single-step correction, or (3) domain-specific verification algorithms that prioritize numerical consistency.

\begin{figure}[t]
  \begin{center}
    \centerline{\includegraphics[width=\columnwidth]{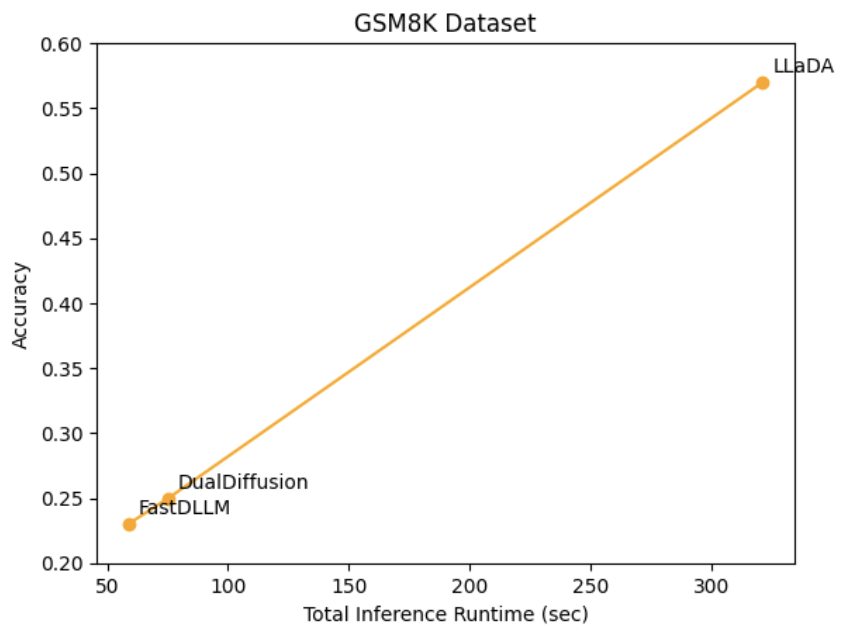}}
    \caption{Accuracy comparison on GSM8K. DualDiffusion underperforms on mathematical reasoning, indicating sensitivity to drafter quality in multi-step tasks.}
    \label{fig:gsm8k_accuracy}
  \end{center}
\end{figure}

Despite the accuracy gap, DualDiffusion maintains a 4.0$\times$ speedup over LLaDA on GSM8K (75.4s vs. 298.3s), demonstrating that the pipeline remains computationally efficient even when verification is less effective. This highlights an important direction for future work: adaptive verification strategies that increase scrutiny on tasks where the drafter is unreliable.

\subsection{Discussion}

DualDiffusion successfully achieves a better Pareto frontier on MMLU, maintaining near-verifier accuracy while delivering substantial speedups. The results demonstrate that speculative decoding is viable for masked diffusion models when the drafter and verifier distributions are sufficiently aligned. However, performance on GSM8K reveals that current verification strategies are insufficient for high-precision, multi-step reasoning tasks.

The discrepancy between MMLU and GSM8K performance highlights a fundamental challenge in MDM speculative decoding: unlike AR models where token-level errors are localized, errors in bidirectional generation can propagate unpredictably through the entire sequence. Future work should explore: (1) adaptive verification that adjusts scrutiny based on task difficulty, (2) multi-step verification rather than single-shot correction, and (3) confidence-calibrated verification that accounts for drafter uncertainty on specific token types (e.g., numerical digits).



\section{Conclusion}

We introduce DualDiffusion, a speculative decoding framework for masked diffusion language models that combines fast drafter models with accurate verifier models to improve the quality-efficiency Pareto frontier. Our approach runs multiple unmasking steps with an accelerated drafter (e.g., FastDLLM or dKV-Cache) and performs selective error correction using a single verifier step, thereby amortizing the computational cost of high-quality generation.

We propose three verification algorithms—Trust Verifier, KL Divergence-based remasking, and Confidence-based remasking—that balance correction accuracy with computational overhead. Experiments on MMLU demonstrate that DualDiffusion achieves near-verifier accuracy (0.47 vs. 0.48) while delivering 3.9$\times$ speedup compared to running LLaDA alone. This confirms that speculative decoding is viable for MDMs when drafter and verifier distributions are sufficiently aligned.

However, results on GSM8K reveal current limitations: DualDiffusion achieves only 0.25 accuracy compared to the verifier's 0.57, indicating that single-step verification is insufficient for multi-step mathematical reasoning where drafter errors cascade through solution paths. This task-dependent performance highlights fundamental challenges in adapting speculative decoding to non-autoregressive generation, where bidirectional context dependencies complicate error localization and correction.

Our work establishes DualDiffusion as an effective acceleration strategy for MDMs on logical reasoning tasks while identifying critical directions for future research. Key areas for improvement include: (1) \textbf{Adaptive verification} that increases scrutiny on high-variance tasks like arithmetic, (2) \textbf{Multi-step verification} where the verifier performs multiple correction passes rather than single-shot remasking, (3) \textbf{Task-aware verification algorithms} that specialize for numerical consistency, sequential reasoning, or other domain-specific constraints, and (4) \textbf{Memory-efficient architectures} such as parameter sharing between drafter and verifier to reduce the current 2$\times$ memory overhead.

More broadly, DualDiffusion demonstrates that the speculative decoding paradigm can extend beyond autoregressive models to non-causal generation frameworks. By developing verification strategies tailored to bidirectional attention and iterative refinement, we show that the quality-efficiency trade-offs inherent in approximate MDM acceleration can be mitigated through strategic combination of fast and accurate models. This opens possibilities for applying similar techniques to other non-autoregressive architectures in machine translation, structured prediction, and controllable generation.

\bibliography{main}

@inproceedings{sahoo2024simple,
  title={Simple and Effective Masked Diffusion Language Models},
  author={Sahoo, Subham Sekhar and Arriola, Marianne and Schiff, Yair and Gokaslan, Aaron and Braverman, Edgar and Kuleshov, Volodymyr and Chiu, Justin T and Rush, Alexander M},
  booktitle={Advances in Neural Information Processing Systems},
  year={2024}
}

@article{yang2025esoteric,
  title={Esoteric Language Models},
  author={Yang, Zhihan and Sahoo, Subham Sekhar and Akhauri, Yash and Liu, Johnna and Singh, Deepansha and Cheng, Zhoujun and Liu, Zhengzhong and Xing, Eric and Thickstun, John and Vahdat, Arash},
  journal={arXiv preprint arXiv:2506.01928},
  year={2025}
}

@article{leviathan2023fast,
  title={Fast inference from transformers via speculative decoding},
  author={Leviathan, Yaniv and Kalman, Matan and Matias, Yossi},
  journal={arXiv preprint arXiv:2211.17192},
  year={2023}
}

@article{chen2023accelerating,
  title={Accelerating large language model decoding with speculative sampling},
  author={Chen, Charlie and Borgeaud, Sebastian and Mensch, Arthur and Sutskever, Ilya and Sifre, Laurent and Vinyals, Oriol and others},
  journal={arXiv preprint arXiv:2302.01318},
  year={2023}
}

@inproceedings{austin2021structured,
  title={Structured denoising diffusion models in discrete state-spaces},
  author={Austin, Jacob and Johnson, Daniel D and Ho, Jonathan and Tarlow, Daniel and Van Den Berg, Rianne},
  booktitle={Advances in Neural Information Processing Systems},
  pages={17981--17993},
  year={2021}
}

@article{wu2025fastdllm,
  title={Fast-dLLM: Training-free Acceleration of Diffusion LLM by Enabling KV Cache and Parallel Decoding},
  author={Wu, Chengyue and Zhang, Hao and Xue, Shuchen and Liu, Zhijian and Diao, Shizhe and Zhu, Ligeng and Luo, Ping and Han, Song and Xie, Enze},
  journal={arXiv preprint arXiv:2505.22618},
  year={2025}
}

@article{ma2025dkvcache,
  title={dKV-Cache: The Cache for Diffusion Language Models},
  author={Ma, Xinyin and Yu, Runpeng and Fang, Gongfan and Wang, Xinchao},
  journal={arXiv preprint arXiv:2505.15781},
  year={2025}
}

@article{hendrycks2021measuring,
  title={Measuring massive multitask language understanding},
  author={Hendrycks, Dan and Burns, Collin and Basart, Steven and Zou, Andy and Mazeika, Mantas and Song, Dawn and Steinhardt, Jacob},
  journal={Proceedings of ICLR},
  year={2021}
}

@article{cobbe2021training,
  title={Training verifiers to solve math word problems},
  author={Cobbe, Karl and Kosaraju, Vineet and Bavarian, Mohammad and Chen, Mark and Jun, Heewoo and Kaiser, Lukasz and Plappert, Matthias and Tworek, Jerry and Hilton, Jacob and Nakano, Reiichiro and others},
  journal={arXiv preprint arXiv:2110.14168},
  year={2021}
}
\bibliographystyle{icml2026}

\end{document}